\documentclass[12pt]{article}
\usepackage{neurips_2024}

\usepackage[utf8]{inputenc} 
\usepackage[T1]{fontenc}    
\usepackage{hyperref}       
\usepackage{url}            
\usepackage{booktabs}       
\usepackage{amsfonts}       
\usepackage{nicefrac}       
\usepackage{microtype}      
\usepackage{xcolor}         
\usepackage{graphicx}
\usepackage{tabularx}
\usepackage{enumitem}
\usepackage[most]{tcolorbox}
\usepackage{natbib}
\usepackage{titlesec}
\usepackage{subcaption}
\usepackage{float}
\usepackage{placeins}
\usepackage{footmisc}
\usepackage{amssymb}
\setlist[enumerate]{left=0pt, itemsep=0pt, topsep=0pt}
\usepackage[font=scriptsize]{caption}
\usepackage{xcolor}
\usepackage{tcolorbox}

\title{Nano Bio-Agents (NBA): \\ Small Language Model Agents for Genomics}

\begin{document}
\maketitle

\begin{abstract}
We investigate the application of Small Language Models (<10 billion parameters) for genomics question answering via agentic framework to address hallucination issues and computational cost challenges. The Nano Bio-Agent (NBA) framework we implemented incorporates task decomposition, tool orchestration, and API access into well-established systems such as NCBI and AlphaGenome. Results show that SLMs combined with such agentic framework can achieve comparable and in many cases superior performance versus existing approaches utilising larger models, with our best model-agent combination achieving 98\% accuracy on the GeneTuring benchmark. Notably, small 3-10B parameter models consistently achieve 85-97\% accuracy while requiring much lower computational resources than conventional approaches. This demonstrates promising potential for efficiency gains, cost savings, and democratization of ML-powered genomics tools while retaining highly robust and accurate performance.
\end{abstract}

\vspace{0.5em}

\section{Introduction}

\subsection{Motivation and Context}
The integration of artificial intelligence in bioinformatics has experienced remarkable growth, with large language models (LLM) demonstrating great potential in biomedical applications including clinical trial matching, drug discovery, and genomics research\cite{sarumi2024large}. However, several critical challenges remain that limit their widespread adoption in bio-informatics applications.

A primary concern is the hallucination problem inherent in auto-regressive LLMs ~\cite{kalai2025languagemodelshallucinate}. Since these models lack intrinsic mechanisms to consult with authoritative sources of scientific knowledge, they frequently generate plausible-sounding but factually incorrect content when answering domain-specific queries~\cite{ji2023survey}. In genomics, where precision and accuracy are paramount, such hallucinations can severely compromise research outcomes and clinical decision-making.

To address these accuracy concerns, researchers have increasingly turned to tool-augmented approaches that combine LLMs with external databases and web APIs. This paradigm shift from relying solely on internal "memorized knowledge" to leveraging tool use enables models to access to both accurate and current genomics data from well-established repositories such as NCBI~\cite{gao2023palprogramaidedlanguagemodels, mialon2023augmented}.

However, a significant barrier to widespread adoption remains: the substantial economic burden imposed by large models on research institutions and individual researchers. Current state-of-the-art approaches typically require expensive, commercial LLMs with hundreds of billions of parameters, creating accessibility challenges for resource-constrained academic environments and limiting democratization of AI-powered genomics tools. This motivates our investigation into whether Small Language Models (SLMs) with fewer than 10 billion parameters can achieve comparable performance through architectural intelligence rather than parameter scaling.

\subsection{Previous Work}
A prime example of successfully leveraging tool-augmented LLMs for genomics Q\&A is GeneGPT ~\cite{jin2024genegpt}. This innovative work demonstrated how in-context learning combined with NCBI Web API integration could achieve state-of-the-art performance on the GeneTuring benchmark~\cite{hou2023geneturing}, dramatically outperforming previous approaches. The method employs a simple but effective prompting technique comprising four key components: (1) task instructions, (2) API documentations providing natural language descriptions of NCBI E-utils~\cite{sayers2019database} and BLAST functionality~\cite{altschul1990basic}, (3) concrete demonstrations showing API usage patterns across different genomics tasks, and (4) the target question. The inference algorithm integrates API execution directly into the model's decoding process using Codex~\cite{chen2021evaluating}, detecting API call indicators (``->'' symbols) and seamlessly incorporating real-time database results into the generation flow.

This approach achieved remarkable results, obtaining an average score of 0.83 on nine GeneTuring tasks, a significant improvement over the previous state-of-the-art (0.44 by Bing). The method's effectiveness stems from its ability to compose complex genomics API queries by integrating instruction, documentation \& demonstration, thereby bridging the gap between natural language questions and structured data retrieval. 

Despite GeneGPT's strong performance, our preliminary analysis reveals a degradation in accuracy when the underlying LLM becomes smaller in size. The original results were produced by Codex (code-davinci-002) which is a fine-tuned version of GPT-3 (175 billion parameters) ~\cite{brown2020language} using code data, suggesting its size is of similar magnitude. As we experimented with smaller models, we found the performance deteriorated (Figure~\ref{fig:method_comparison}), resulting from irregular behaviour such as incorrect URL construction or failure to parse the returned document. The accuracy scores decreased significantly and can vary widely, especially when we use SLMs as the underlying model.

\subsection{Research Question \& Contributions}
We consider the following \textbf{Core Research Questions}:

\begin{itemize}[label={$\blacktriangleright$}, leftmargin=0.5cm]
\item Can SLMs achieve large model performance in genomics Q\&A through architectural refinement involving agents?
\item Can the same accuracy be maintained across model size, model architecture and model family?
\item Can the approach be generalized to other applications and tasks?
\end{itemize}

Specifically, we investigate whether an agentic framework that decomposes complex genomics queries into specialized sub-tasks can enable SLMs with fewer than 10 billion parameters to match or exceed the accuracy of models requiring 150+ billion parameters. This research question addresses a fundamental tension in AI-powered scientific applications: the trade-off between model capability and computational accessibility. While previous work has demonstrated that tool augmentation can enhance LLM performance in genomics, the persistent requirement for large, expensive models creates a significant barrier to the democratization of these technologies.

\textbf{Main Contributions:} This work contributes to the intersection of small language models and genomics applications as follows:

\begin{enumerate}
\item \textbf{Agentic Framework Design:} We applied an agentic architecture that enables SLMs to achieve 85-97\% accuracy on the GeneTuring benchmark, exceeding the 83\% reported recently by alternative approaches utilising larger language models. Our framework demonstrates that architectural intelligence can overcome parameter size limitations through strategic task decomposition and tool orchestration.

\item \textbf{Cross-Model Robustness:} We provide a comprehensive evaluation across 50 language models ranging from 1B to 1T+ parameters, demonstrating that our approach maintains robust performance across diverse model families and architectures. This extensive validation establishes the generalizability of our architectural approach beyond specific model implementations.

\item \textbf{Significant Efficiency Gains:} By enabling competitive performance with 7-10B parameter models, our approach realizes the 10-30× efficiency gains predicted for SLM deployments \citep{belcak2025small}. These gains directly address accessibility barriers that limit adoption of AI-powered genomics tools in resource-constrained environments.

\item \textbf{Extensibility and Generalization:} We demonstrate the framework's extensibility through integration with cutting-edge tools such as AlphaGenome, showcasing how the NBA architecture can adapt to incorporate emerging genomics technologies. This positions our approach as a sustainable foundation for future developments in AI-powered genomics research.

\item \textbf{Security and Privacy Enhancement:} Our SLM-centric framework enables local inference deployment in production environments, ensuring that sensitive genomics data and research queries remain within institutional boundaries without transmission to external commercial APIs. This addresses critical privacy concerns in individual clinical setting and proprietary research environments while maintaining full functionality through locally executable SLMs.
\end{enumerate}

These contributions collectively demonstrate that architectural improvement can democratize access to sophisticated genomics AI capabilities, enabling broader adoption across academic institutions, clinical environments, and individual researchers who previously lacked access to large-scale computational resources.

\subsection{Paper Organization}
The paper is organized as follows. Section~\ref{sec:background} provides background on genomics question answering, reviews related work on tool-augmented LLMs, and discusses the emergence of small language models in agentic systems. Section~\ref{sec:methodology} presents the NBA framework architecture, compares it with GeneGPT's approach, and details our experimental design. Section~\ref{sec:results} reports comprehensive performance results across a wide spectrum of model sizes and families to illustrate the resilience and efficiency gains of the approach. Section~\ref{sec:discussion} discusses key insights, implications for democratizing genomics AI, and limitations of our approach. Finally, Section~\ref{sec:conclusion} summarizes our contributions and outlines future research directions.

\section{Background \& Related Work}\label{sec:background}

\subsection{Genomics Question Answering}
Following ~\cite{jin2024genegpt}, we focus on genomics question answering in the GeneTuring benchmark dataset~\cite{hou2023geneturing} comprised of 450 questions across 9 categories. They fall into the following areas:

\textbf{Nomenclature:} Gene alias and gene name conversion tasks that require finding official gene symbols for non-official synonyms.

\textbf{Genomic Location:} Tasks involving gene location, SNP location, and gene-SNP association, where models must identify chromosome locations (e.g., "chr2") or determine relationships between genetic variants and genes.

\textbf{Functional Analysis:} Gene-disease association tasks that identify disease-related genes, and protein-coding gene classification that determines whether genes encode proteins.

\textbf{Sequence Alignment:} DNA sequence alignment tasks that map sequences to human chromosomes or identify species origins using BLAST-based approaches.

Each task contains 50 question-answer pairs and requires precise access to NCBI databases, making them ideal benchmarks for evaluating tool-augmented language models in genomics applications. Finally, the short, definitive answer enable clear scoring and measurement of success, thus enabling comparison across models and approaches.

\subsection{LLM Approaches in Genomics}
The application of large language models to genomics has emerged as a rapidly growing field, with comprehensive reviews covering the breadth of LLM applications in bioinformatics, such as ~\cite{lin2025bridging}. In the specific context of genomics question answering, the simplest approach involves directly querying LLMs with genomics questions, relying solely on their pre-trained knowledge. While straightforward to implement, this method suffers from significant hallucination issues, achieving sub-optimal results on GeneTuring tasks due to the models' inability to access current, authoritative genomics databases. Tool-augmented methods like GeneGPT~\cite{jin2024genegpt} integrated NCBI API calls with LLM reasoning, demonstrating that external tool access can dramatically improve accuracy. This approach achieved SOTA (State-Of-The-Art) performance with 83\% accuracy by combining few-shot in-context learning with real-time database queries, effectively bridging the gap between natural language understanding and structured genomics data. Other approaches like BioMaster~\cite{su2025biomaster} and GeneAgent~\cite{wang2025geneagent} have since emerged to explore the application of agentic systems to tackle the challenge.

\subsection{Small Language Models in Scientific Applications}
Recent research has increasingly recognized the potential of Small Language Models (SLMs) as efficient alternatives to large-scale models, particularly in specialized domains ~\cite{subramanian2025smalllanguagemodelsslms, garg2025risesmalllanguagemodels}. The NVIDIA perspective articulated in ~\cite{belcak2025small} argues that SLMs offer compelling advantages for agentic applications through architectural intelligence rather than parameter scaling.

\textbf{Key Benefits of SLMs:} Cost efficiency represents a primary advantage, with SLMs achieving 10-30× reduction in FLOPs and latency compared to large models (~\cite{belcak2025small}). Deployment flexibility enables edge computing and local inference, crucial for resource-constrained environments. Additionally, SLMs provide enhanced transparency and debuggability, making them more suitable for scientific applications where interpretability is paramount. Environmental sustainability concerns also favor SLMs due to their significantly reduced computational footprint.

\textbf{Capability Evidence:} Modern SLMs have demonstrated the ability to match larger models on specialized tasks through strategic design and domain-specific optimization. Examples include numerous open model families mentioned in ~\cite{belcak2025small} that can achieve comparable performance to 30B+ models while running faster and at a fraction of the computational cost.

These developments suggest that the key to effective AI deployment lies not in model size but in architectural innovation tailored to specific application domains—a principle that directly motivates our investigation into SLM-based genomics applications.

\subsection{Agentic Architectures}
The rapid adoption of agentic AI systems across diverse domains—from software development and scientific research to business automation—reflects a fundamental shift toward modular, tool-augmented approaches that leverage specialized capabilities rather than relying solely on monolithic model scaling. Recent surveys highlight the particular promise of agentic systems in scientific discovery, where complex, multi-step tasks require coordination between different tools and reasoning components~\cite{gridach2025agentic}. Modern frameworks like LangChain LCEL (LangChain Expression Language) provide standardized approaches for building agentic systems, enabling modular design patterns that facilitate development, testing, and deployment of complex AI workflows.

\section{Methodology}\label{sec:methodology}

\subsection{NBA Framework Architecture}\label{sec:nba_architecture}
We utilize a simple and effective agentic framework for genomics question answering that decomposes complex queries into modular sub-tasks. The NBA framework consists of core components below implemented in a sequential pipeline:

\begin{figure}[h]
\centering
\includegraphics[width=0.6\textwidth]{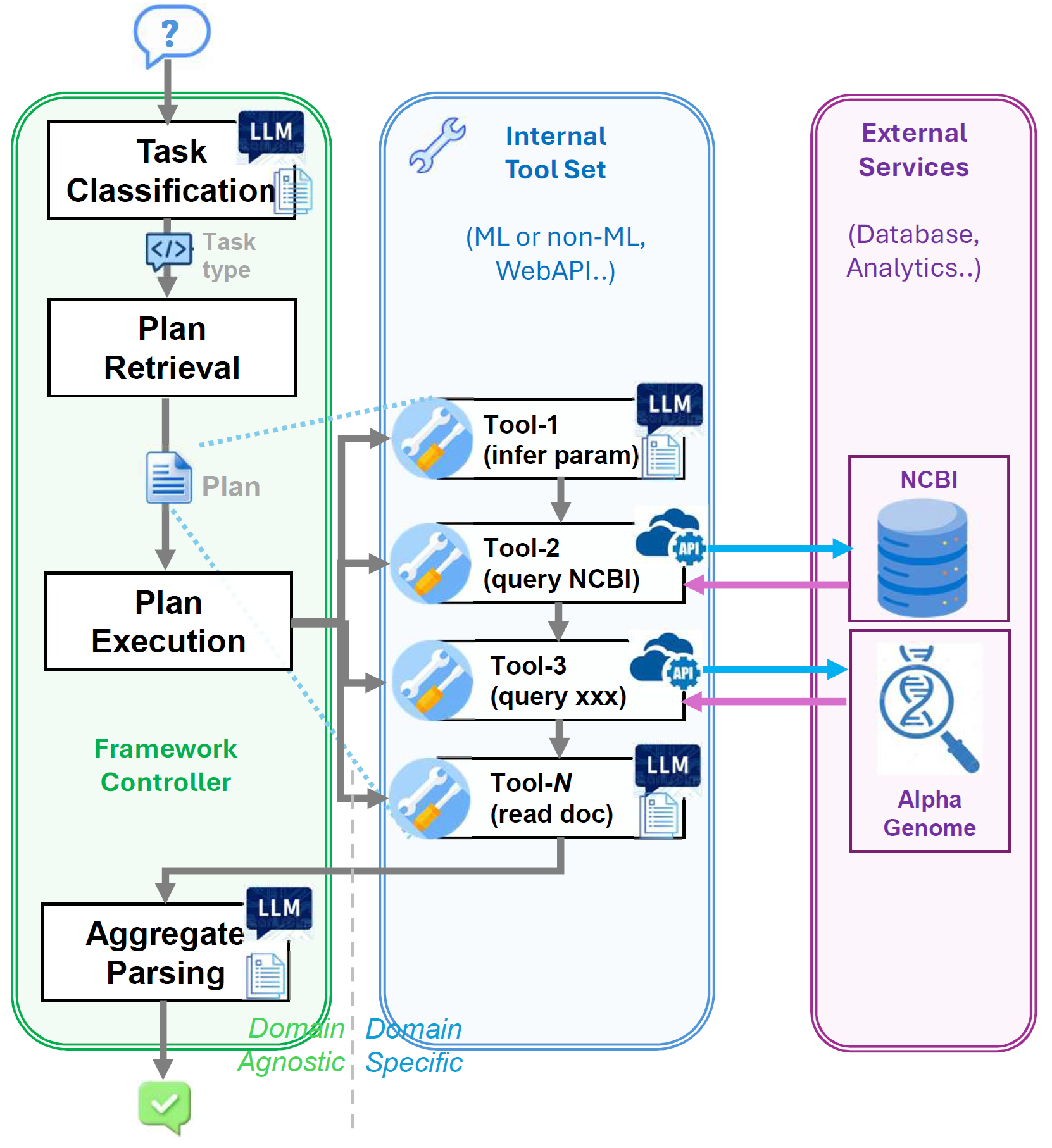}
\caption{NBA Framework design and implementation.}
\label{fig:nba_diagram}
\end{figure}

\begin{itemize}
\item \textbf{Task Classification:} Using LLM-based in-context learning to identify the type of genomics query (e.g., gene nomenclature, genomic location, functional analysis, or sequence alignment) from the input question.

\item \textbf{Plan Retrieval:} Extract appropriate execution templates from a curated dataset based on the classified task type, providing structured guidance for subsequent processing steps. Each plan consists of the steps needed and are encoded in functions with well-defined inputs and outputs.

\item \textbf{Plan Execution invoking Tools :} Calling individual tools through coded function implementations, either involing S/LLM or not. Examples include: 
\begin{itemize}
\item \textbf{NCBI Web API calls} (E-utils, BLAST) that retrieve authoritative genomics data.
\item\textbf{Input Parameter Inference:} LLM-based extraction of specific parameters (gene names, sequences, disease terms, etc.) from the natural language query using in-context learning with domain-specific examples.
\item \textbf{Document Parsing} LLM-based extraction of relevant information from the raw API responses (a Specialist-role).
\end{itemize}

\item \textbf{Aggregate Result Parsing:} LLM-based interpretation and final aggregation (a Generalist-role), formatting the output into user-friendly, readable answers.
\end{itemize}

\textbf{Design Rationale:} The motivation stems from the observation that delegating the entire chain of operations to an LLM through an overarching prompt (as in GeneGPT) places excessive burden on the model's ability to generalize from few-shot examples and follow instructions precisely. By decomposing the process into individual tasks and applying different techniques (LLM reasoning or coded functions) as appropriate, we leverage each component's strengths while ensuring more robust execution.

This modular approach also enables better monitoring, debugging, and diagnosis by pinpointing exactly where failures occur in the pipeline. Instead of a single monolithic in-context prompting, this "Divide and Conquer“ approach corresponds to the "Code Agency" mode in the ~\cite{belcak2025small} paper and takes advantage of both the robustness of the coded implementation of Web API or Analytics calls and the powerful parsing and language understanding capabilities offered by LLMs.

Finally, we chose a simple linear execution path as it proves sufficient for the genomics question answering tasks at hand. This design can naturally be generalized to more complex directed acyclic graph (DAG) structures for future applications requiring more complex reasoning workflows. 

\subsection{Technical Implementation}
The NBA framework is built with the intention of being generic and domain-agnostic, with clear code separation between the core agentic architecture and domain-specific knowledge.

\textbf{Generic Framework with Dynamic Content:} The agent framework contains no coded domain knowledge from bioinformatics. Instead, domain expertise is captured in other dedicated modules and JSON configuration files, ensuring no leakage between architectural logic and genomics-specific content. SLMs learn from examples and instructions through in-context learning, but within a much narrower scope compared to monolithic approaches like GeneGPT, reducing the cognitive burden on individual model calls.

\textbf{Optimized Functional Implementation:} API calls and other programmatic functions are implemented once and optimized with a caching mechanisms (handling the double BLAST queries needed for analysis wait-time). This approach eliminates redundant database queries and significantly improves response times for frequently queried data.

\textbf{Pure Query Function Implementation:} In addition to the LLM-based pipeline, we implemented pure query functions that bypass language models entirely for the 9 GeneTuring task categories, invoked by setting "method" to "code" with the same API. The functions first compute cosine similarity between input question embeddings and stored question embeddings to determine the task types (subject to thresholds) and then work out the arguments for URL construction, enabling direct NCBI API calls without LLM intervention for exact or near-exact matches. Their deterministic implementations provide reliable and useful benchmark results so that one can generate additional Q\&A pairs with minimal manual overhead for testing and validation.

\textbf{Framework Infrastructure:} The system leverages standard model API's and LangChain LCEL (LangChain Expression Language) as a modular, configuration-driven framework that enables flexible pipeline construction. The wide adoption of LangChain and standard API (e.g., OpenAI) means that it takes little effort to invoke new LLMs offered by the popular inference provider platforms. We can therefore conduct comprehensive testing of 80+ models across model and inference providers (OpenAI, Anthropic, Google, Qwen, Mistral, Microsoft, IBM, Meta, NVIDIA NIM, HuggingFace, and local Ollama deployments). Other technical features we put in include detailed logging (e.g., timing, memory usage, and token consumption), smart truncation for efficiency, comprehensive retry logic, and failure recovery mechanisms.

\subsection{Experimental Design}
We evaluate the effectiveness of the Agentic framework method using the scoring methodology and dataset similar to those in ~\cite{jin2024genegpt} with minor updates (see Appendix \ref{sec:appendix_refinement_dataset} for details). Specifically, we compute the average scores for each of the 9 tasks in the benchmark GeneTuring dataset as well as the overall total average, ranged from 0 to 100\%. We tested the method against the Direct Prompting of the LLM and the GeneGPT across a wide range of open and proprietary models with different parameter counts and model architecture (e.g. Dense and MoE). We also ran the pure query function mode to ensure the answers match the stored answers.

For the other methods in our comparative analysis, care had to be taken to construct meaningful results. For example, in the Direct prompting approach, we have to instruct the LLM to not use external tools or websearch, for otherwise it could find the answers on the GitHub repository (e.g. the public GeneGPT code base) or paper in ArXiv (e.g., the GeneGPT paper) that contain exactly the question-and-answer pair.

In addition, we estimate the token counts, both the input consumptions and output generated, by using a simple conversion formula from the number of input and output characters. The ratio of the conversion is calibrated using the sample queried NCBI documents from the web query to ensure the approximation is appropriate. The token counts are then translated into dollar estimates of the economic costs incurred for each approach, using the publicly available information of API costs for each inference provider for each model. Whilst this is only an approximation of the actual costs, the aim is to gauge the relative (rather than absolute) efficiency saves of the method against other approaches. Finally, we captured the elapsed time and memory consumption too. All of these metrics are logged at the question level, and in many cases down to the sub-task level, for detailed analysis.

\section{Results}\label{sec:results}

\subsection{Performance Across Model Scales}\label{sec:performance_robustness}
\begin{figure}[htbp]
\centering
\includegraphics[width=\textwidth]{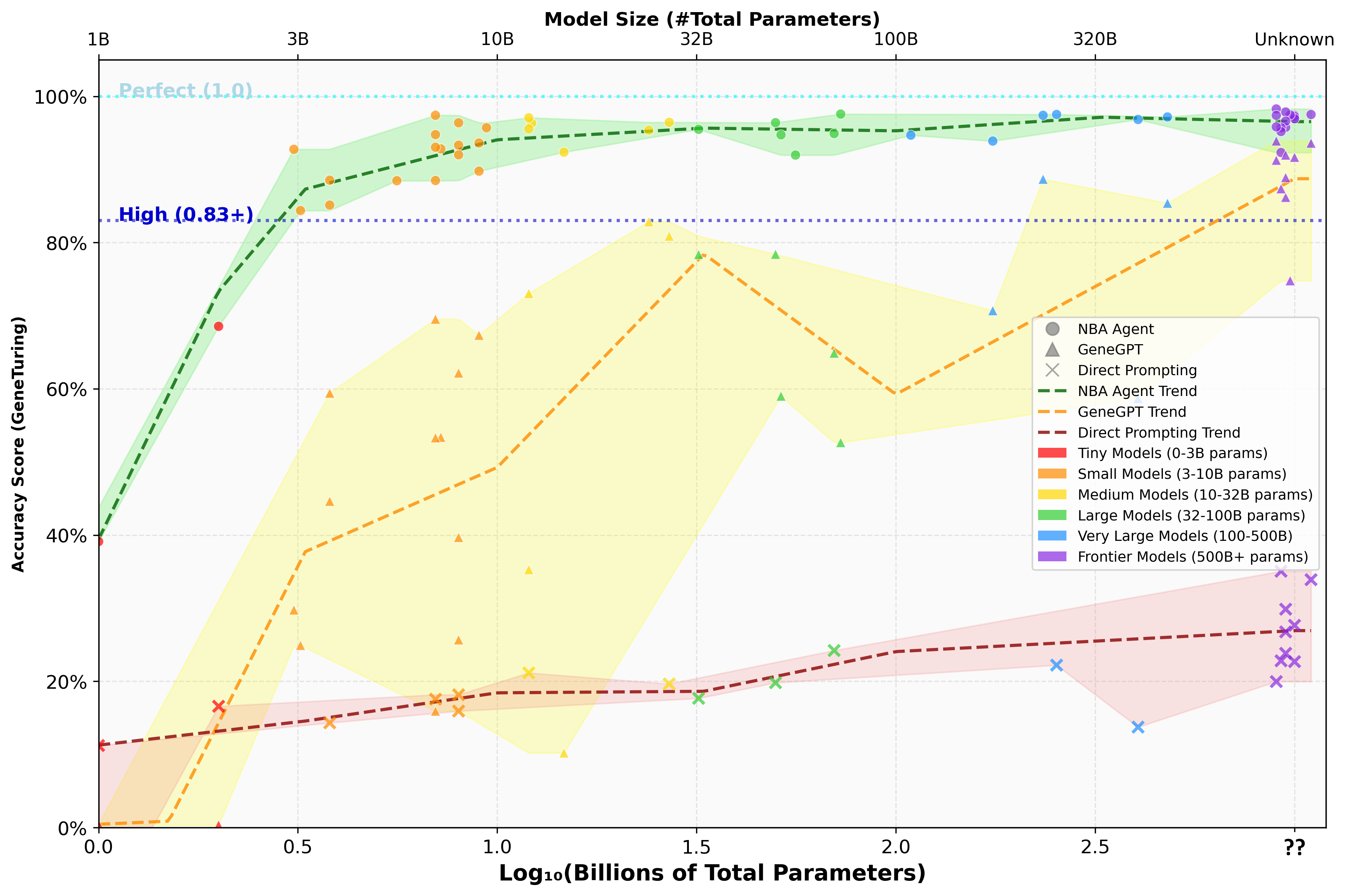}
\caption{Performance comparison across three approaches: Direct prompting, GeneGPT, and NBA agentic framework. Accuracy (average score on GeneTuring dataset) versus Model Parameter counts (log-scale). Most flagship models (current or past) do not have the parameter information public so we place them in one bucket on the right-hand side. The dash line indicating "High" performance is somewhat arbitrary and chosen as such because the previous SOTA in GeneGPT~\cite{jin2024genegpt} was reported at this level.}
\label{fig:method_comparison}
\end{figure}

We evaluated NBA on 50 models with parameter counts spanning several orders of magnitude to assess its performance consistency across different model sizes. Figure~\ref{fig:method_comparison} shows the relationship between model parameter count and accuracy score on the GeneTuring dataset for NBA, GeneGPT, and direct LLM prompting. Note that many commercial model owners do not disclose their parameter information publicly. We therefore place their experiment results all under one bucket ("Unknown") on the far right hand side of the plot. Consequently we request the readers to interpret the plot in those region only qualitatively and not draw quantitative conclusion regarding its performance gradient in those regions. Finally, for MoE models, we use Total number of parameters rather than Active parameter counts. This way of presentation leans on the conservative side in terms of interpreting the results (e.g., we did not want to classify a potentially good result by a model with many experts (each of which is small) to be in the same category of a non-MoE SLM).

\FloatBarrier 
\FloatBarrier

For the models with public information on their parameter counts, we observe in Figure~\ref{fig:method_comparison} that the NBA agentic approach shows consistently strong performance, from 1 trillion all the way down to the 3 billion parameters range. Models in the 7-10B parameter range consistently achieve 88\% to 97\% accuracy and the performance extend across the model size and variety of model families. This suggegsts that we can, with a decent level of confidence, utilise most models, SLM or LLM, Open or Proprietary, dense or MoE, provided the parameter counts are above the 3 billion mark. The "sweet spot" of 7-10B is particularly noteworthy given it simultaneously achieves accuracy and cost efficiency.\\

As expected, the accuracy deteriorates sharply as we reach the 1 to 2 Billion region, where the model loses its ability to follow the instruction reliably. Note that, however, the worst performance of the 1B model tested here is still higher than the result from Directly prompting any LLM, even for very large models. Indeed, as we inspect the performance of the direct LLM prompting approach, we observe a consistent low score due to its hallucination nature, consistent with the previous findings in literature. For example, in tasks involving DNA sequence alignments, the direct LLM prompting approach is purely guessing given it lacks the access to the genenomic database and thus achieves very low scores. \\

The other tool-augmented approach, GeneGPT, exhibits strong performance when we used large flagship models such as the OpenAI's GPT, Anthropic's Claude and Google's Gemini series. However, as we swap in smaller models with lower parameter counts, the accuracy decreases and we saw much bigger variation in the scores: some models retain certain level of accuracy whilst others may lose it rapidly. This The resulting performance variability could undermine the reliability needed for deploying smaller, open-source models in production environments. 

\begin{figure}[h]
\centering
\includegraphics[width=1.0\textwidth]{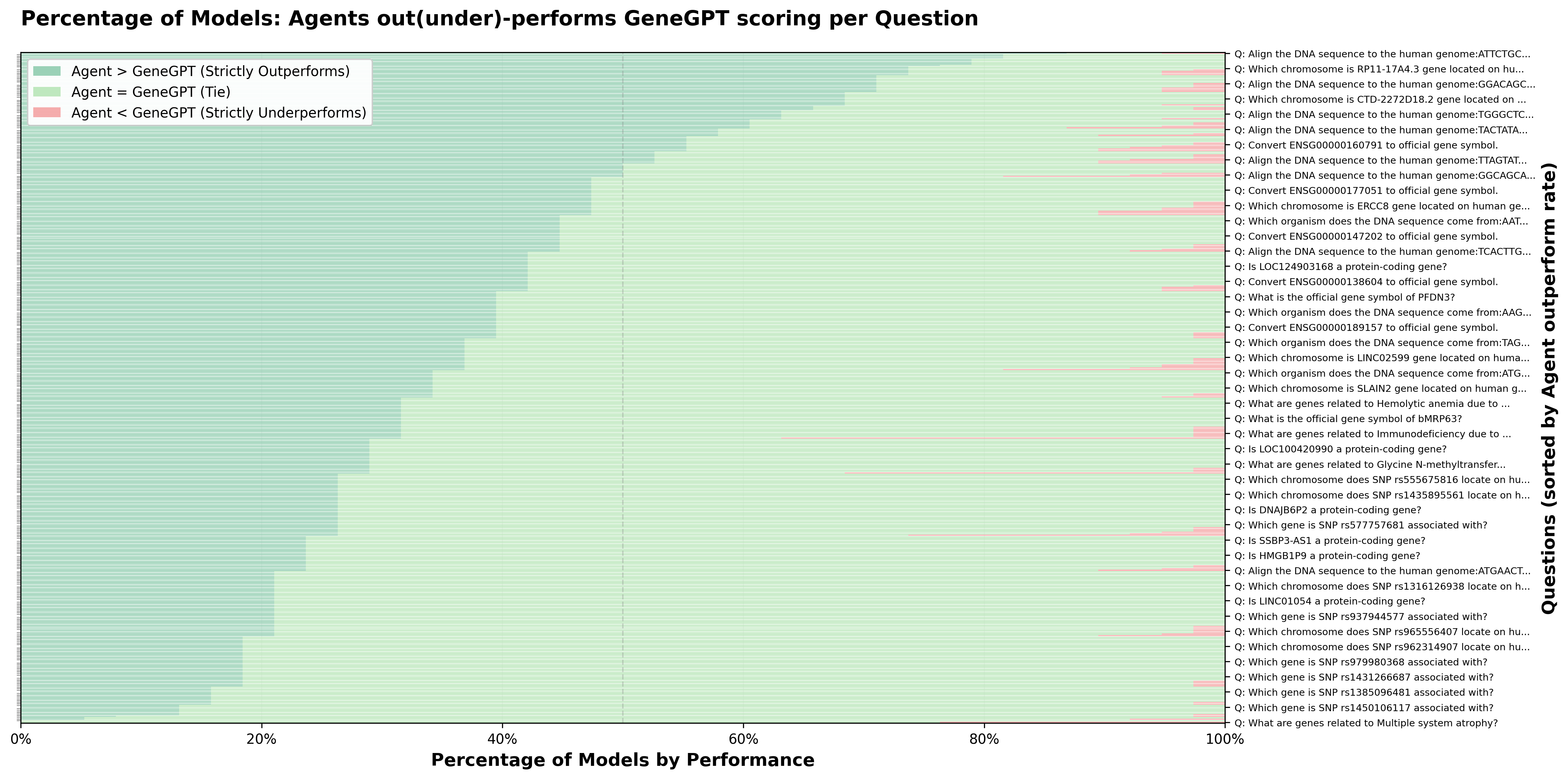}
\caption{Distribution of scores per question across different models and approaches, showing consistency patterns in performance. (Results for all questions are plotted and due to spacing only a sample of the question texts are shown on the vertical axis.)}
\label{fig:scores_per_question}
\end{figure}

The comparison above in Figure ~\ref{fig:method_comparison} is using the top-level scores averaged across all the questions in the dataset. As a follow-up we dived into the comparison on a per-question basis and investigate whether the NBA agents could sometimes underperform other methods and, if so, whether there is any pattern. We plot the frequencies of whether any of the models "favors" the Agentic or the GeneGPT approach for every question in the dataset. Specifically, we compute the percentage of the models where Agnetic's score is higher, equal or lower to the GeneGPT approach for each question. Figure ~\ref{fig:scores_per_question} shows that the Agentic approach almost always perform either better or at-par in comparison, and only in very rare occasion that it did not work as well for certain specific models. \\

It is worth noting that NBA was specifically designed with architectural optimizations for smaller models, while we applied GeneGPT's original methodology without modification or refinement. The difference in design focus may contribute to the observed performance patterns, even though both approaches utilize the same underlying genomics databases and APIs. \\

In summary, NBA demonstrates consistent and strong performance across model sizes (from 1T+ down to 3\~4 B size) compared to alternative approaches, as evidenced by high accuracies, low variance, and great resilience across model choices.

\subsection{Performance Across Model Families}\label{sec:cross_model}
\vspace{-0.2cm}
\begin{table}[H]
\centering
\footnotesize
\begin{tabular}{clrc}
\toprule
\textbf{Family} & \textbf{Model} & \textbf{Score} & \textbf{Size (B)} \\
\midrule
\midrule
Claude & claude-opus-4-1-20250805 & 97.51\% & \textcolor[gray]{0.7}{??} \\
Claude & claude-sonnet-4-20250514 & 97.85\% & \textcolor[gray]{0.7}{??} \\
Claude & claude-3-7-sonnet-20250219 & 95.69\% & \textcolor[gray]{0.7}{??} \\
Claude & claude-3-5-haiku-20241022 & 98.26\% & \textcolor[gray]{0.7}{??} \\
\midrule
GPT & gpt-4.1 & 97.39\% & \textcolor[gray]{0.7}{??} \\
GPT & gpt-4o & 97.22\% & \textcolor[gray]{0.7}{??} \\
GPT & gpt-4.1-mini & 96.60\% & \textcolor[gray]{0.7}{??} \\
GPT & gpt-4.1-nano & 92.31\% & \textcolor[gray]{0.7}{??} \\
GPT & gpt-5-nano & 95.18\% & \textcolor[gray]{0.7}{??} \\
GPT & gpt-4o-mini & 95.82\% & \textcolor[gray]{0.7}{??} \\
GPT & gpt-3.5-turbo & 93.88\% & 175.0 \\
\midrule
Gemini & gemini-2.5-flash-lite & 97.46\% & \textcolor[gray]{0.7}{??} \\
Gemini & gemini-2.0-flash & 95.75\% & \textcolor[gray]{0.7}{??} \\
Gemini & gemini-1.5-flash & 97.39\% & \textcolor[gray]{0.7}{??} \\
Gemini & gemini-1.5-flash-8b & 96.37\% & 8.0 \\
\midrule
Gemma & google/gemma-3-27b-it & 96.43\% & 27.0 \\
Gemma & google/gemma-3-12b-it & 96.32\% & 12.2 \\
Gemma & google/gemma-2-9b-it & 89.76\% & 9.0 \\
Gemma & google/gemma-2-2b-it & 68.54\% & 2.0 \\
Gemma & google/gemma-3-1b-it & 39.12\% & 1.0 \\
\midrule
Llama & meta/llama-3.1-405b-instruct & 96.83\% & 405.0 \\
Llama & meta/llama-4-scout-17b-16e-instruct & 94.68\% & 109.0 \\
Llama & meta/llama-3.3-70b-instruct & 94.90\% & 70.0 \\
Llama & meta/llama-3.1-8b-instruct & 93.31\% & 8.0 \\
Llama & meta/llama-3.2-3b-instruct & 84.38\% & 3.2 \\
\midrule
Mistral & mistralai/mixtral-8x7b-instruct-v0.1 & 91.96\% & 56.0 \\
Mistral & mistralai/mistral-small-3.1-24b-instruct-2503 & 95.36\% & 24.0 \\
Mistral & mistralai/mistral-nemotron & 97.05\% & 12.0 \\
Mistral & mistralai/mistral-7b-instruct-v0.3 & 93.03\% & 7.0 \\
\midrule
Nemotron & nvidia/llama-3.1-nemotron-ultra-253b-v1 & 97.51\% & 253.0 \\
Nemotron & nvidia/llama-3.1-nemotron-51b-instruct & 94.73\% & 51.5 \\
Nemotron & nvidia/llama-3.3-nemotron-super-49b-v1.5 & 96.38\% & 49.9 \\
Nemotron & nv-mistralai/mistral-nemo-12b-instruct & 95.53\% & 12.0 \\
Nemotron & nvidia/nvidia-nemotron-nano-9b-v2 & 93.61\% & 9.0 \\
\midrule
Phi & microsoft/phi-4 & 92.36\% & 14.7 \\
Phi & microsoft/phi-3-small-128k-instruct & 88.48\% & 7.0 \\
Phi & microsoft/phi-4-multimodal-instruct & 88.46\% & 5.6 \\
Phi & microsoft/phi-4-mini-instruct & 88.53\% & 3.8 \\
Phi & microsoft/phi-3.5-mini-instruct & 85.12\% & 3.8 \\
\midrule
Qwen & qwen/qwen3-coder-480b-a35b-instruct & 97.17\% & 480.0 \\
Qwen & qwen/qwen3-235b-a22b & 97.39\% & 234.0 \\
Qwen & qwen/qwen2.5-72B-Instruct & 97.56\% & 72.7 \\
Qwen & qwen/qwen2.5-coder-32b-instruct & 95.48\% & 32.0 \\
Qwen & qwen/qwen2.5-coder-7b-instruct & 97.39\% & 7.0 \\
Qwen & qwen/qwen2.5-7b-instruct & 94.76\% & 7.0 \\
Qwen & qwen/qwen2.5-coder-3B-Instruct & 92.74\% & 3.1 \\
\midrule
Kimi & moonshotai/Kimi-K2-Instruct-0905 & 96.94\% & 1000.0 \\
\hline
GLM & zai-org/glm-4-9b-0414 & 95.69\% & 9.4 \\
\hline
Granite & ibm/granite-3.3-8b-instruct & 91.99\% & 8.0 \\
\hline
Falcon3 & tiiuae/falcon3-7b-instruct & 92.81\% & 7.2 \\
\bottomrule
\end{tabular}
\centering
\captionsetup{justification=centering}
\caption{NBA framework Results by Family across 50 tested models, loosely ordered by Total parameter count (activated or not) or estimates, within family.}
\label{tab:performance_model_families}
\end{table}
\vspace{-0.2cm}
We report scores of the NBA approach across major model families to assess robustness and generalizability beyond specific architectures in Table~\ref{tab:performance_model_families}. Performance remains consistent across different model families, suggesting that NBA's effectiveness is not tied to specific training approaches or architectures. Open-source models perform comparably to proprietary alternatives in our tests: a positive implication for accessibility of genomics AI tools.
\subsection{Performance Across Task Types}\label{sec:task_specific}
\begin{figure}[h]
\centering
\includegraphics[width=1.0\textwidth]{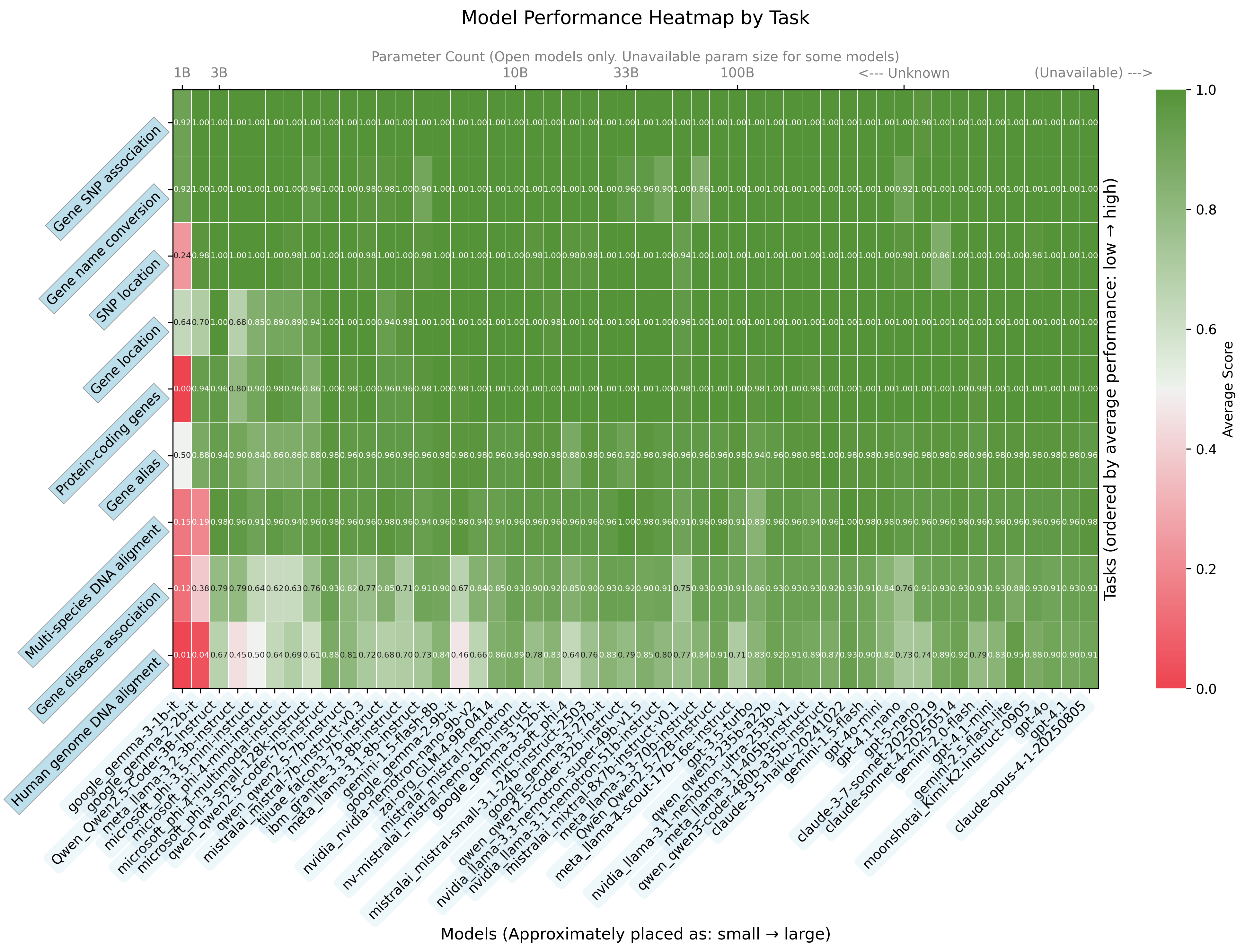}
\caption{NBA performance breakdown by GeneTuring task categories, showing robustness across different genomics domains.}
\label{fig:task_heatmap}
\end{figure}

To investigate whether the SLM agentic approach performs consistently across different question types, we use a comprehensive heatmap Figure~\ref{fig:task_heatmap} to highlight the score variation across all 9 GeneTuring tasks in the 4 primary genomics domains (Nomenclature, Genomic Location, Functional Analysis, and Sequence Alignment) organized by model sizes.

Firstly, we observe that NBA maintains robust performance across all genomics domains, with no single task category showing systematic degradation when using smaller models above 3 billion parameters. Second, certain tasks (notably gene nomenclature and genomic location) show particularly strong performance with SLMs, achieving 80\%+ accuracy consistently across the targeted parameter range. Third, even the most challenging tasks (sequence alignment to multiple species) maintain viable performance levels with appropriately sized SLMs, i.e. there does not exist a performance gap that is hidden or "rescued" by high scores by other tasks through averaging effect.

\section{Discussion} \label{sec:discussion} 

\subsection{SLM and Agentic AI}

Our findings provide strong empirical support for the three-view (3V) thesis articulated in Nvidia's paper, "Small Language Models are the Future of Agentic AI"~\cite{belcak2025small}, demonstrating these principles in the specific context of genomics applications:

\textbf{V1 - Sufficient Capability:} 
\begin{itemize}
\item \textit{"SLMs are principally sufficiently powerful to handle the language modeling tasks required in agentic applications"}. 
\item Our results show that model agents with 7-10B parameters reach 85-97\% accuracy on GeneTuring tasks, demonstrating that specialized reasoning capabilities do not require massive models when properly orchestrated within an agentic framework.
\end{itemize}
\textbf{V2 - Operational Suitability:} 
\begin{itemize}
\item \textit{"SLMs are inherently more operationally suitable for use in agentic systems than LLMs"}. 
\item The modular nature of the NBA framework benefits from SLMs' faster inference times, reduced memory footprint, and enhanced debuggability. The ability to rapidly iterate through task classification, parameter inference, and result parsing steps is crucial for responsive agentic workflows, where multiple model calls are coordinated within a single query resolution.
\end{itemize}

\textbf{V3 - Economic Necessity:} 
\begin{itemize}
\item \textit{"SLMs are necessarily more economical for the vast majority of language model uses in agentic systems than their general-purpose LLM counterparts by virtue of their smaller size"}. 
\item Our analysis supports the claim that a 10-30× efficiency gains in inference costs and energy consumption is achievable, thus making genomics AI accessible to resource-constrained research environments that previously could not afford large model deployments.
\end{itemize}

The genomics domain thus serves as a compelling proof-of-concept for the broader applicability of SLM-based agentic systems across scientific applications, validating the thesis above through concrete performance metrics.

\subsection{Generalization and Extensions}
We experimented with extending the framework to tackle other tasks beyond the Q\&A analysed above. The goal is to see if we can call other tools beyond the NCBI query utilities by following simple pre-defined steps. One candidate we were excited to explore is AlphaGenome ~\cite{alphagenome}, a unifying model by Google DeepMind for deciphering the regulatory code within DNA sequences. The model's analysis ability of DNA sequences of up to 1 million base pairs can be accessed using a public API, making it simple to integrate. We have defined and implemented three prototype tasks that integrate Agentic framework, SLM, NCBI and AlphaGenome:
\begin{enumerate}
\item \textbf{ISM sequence optimization}: e.g. \textit{“Find critical bases in this enhancer DNA sequence: ATCGATC...”}
\item \textbf{SNP splicing analysis}: e.g. \textit{“What splicing impact does SNP rs7903146 have?”}
\item \textbf{Multi-tissue expression}: e.g. \textit{“Compare TP53 expression between heart, liver, and lung tissues”}
\end{enumerate}

The integration of AlphaGenome within our SLM-based agentic framework demonstrates the operational advantages of distributed intelligence across specialized components. The genomics tasks require coordination between multiple heterogeneous services: NCBI E-utils for gene and variant database retrieval, AlphaGenome's REST API for functional predictions, and statistical analysis modules for biological interpretation. It leverages AlphaGenome's Python SDK to access three primary prediction methods: \texttt{predict\_sequence()} for direct DNA analysis, \texttt{predict\_interval()} for genomic coordinate-based queries, and \texttt{score\_ism\_variants()} for systematic mutagenesis. 

\subsubsection{Task-Specific Implementation Analysis}

\textbf{ISM Sequence Optimization:} The framework automates systematic mutagenesis by coordinating hundreds of individual predictions through a single AlphaGenome API call to \texttt{score\_ism\_variants()}. For example, the system selects DNase-seq output with K562 cell line specificity for chromatin accessibility analysis, automatically generates sequence logos using matplotlib integration, and provides transcription factor binding site annotations. Processing a 336 bp enhancer sequence completes in under 30 seconds, representing a significant improvement over manual approaches that typically require hours of manual coordination.

\textbf{SNP Splicing Analysis:} This workflow exemplifies multi-service orchestration, beginning with E-utils queries to the SNP database for variant coordinate retrieval, followed by AlphaGenome's \texttt{predict\_variant()} method for reference versus alternate allele comparison. The framework automatically requests splice junction predictions and quantifies splicing efficiency changes, linking results to clinical databases for disease association context. For example, the rs7903146 analysis demonstrates integration between database retrieval (genomic coordinates: chr10:114,758,349, T→C) and mechanistic prediction (50\% reduction in splice donor usage), completed in under 10 seconds.

\textbf{Multi-Tissue Expression Comparison:} This complex workflow coordinates sequential \texttt{predict\_interval()} calls across multiple tissue ontologies, requiring intelligent tissue-to-ontology mapping (e.g., "liver" → "UBERON:0001114"), statistical analysis across expression predictions, and biological interpretation synthesis. For example, the TP53 analysis across heart, liver, and lung tissues (processing time: 8.8 seconds) demonstrates the framework's capability to handle parallel API coordination while maintaining data provenance and providing standardized statistical metrics that would typically require manual statistical software usage.

\subsubsection{Efficiency Gains}

Initial assessment suggests substantial efficiency gains of the approach compared to traditional manual approaches. The ISM analysis processes 768 mutations (256 positions × 3 alternative bases per position) in 30 seconds versus potentially hours for equivalent manual API coordination and visualization generation. Multi-tissue expression analysis successfully obtained three tissue predictions with total processing time less than 10 seconds, compared to traditional workflows requiring non-trivial efforts for ontology lookup, multiple API calls, and statistical analysis coordination. As we do not have established benchmarks available to measure either the accuracy or the workload, it is premature to quantitatively assess the effectiveness of the approach. Nevertheless we are hopeful that many researchers in the field will be able to advance the enquiry in the very near future. 

\subsection{Limitations and Challenges}
Whilst the investigation so far indicates promising potential for the agentic approach, we are mindful of its limitations. Firstly the tasks we have explored are simple in nature and are solvable using well-defined routines. Many tasks encountered in real-world are much less clear and may require multiple paths of exploration where results are aggregated for the final response. The errors or loss of precision can easily accumulate to impact the quality of the final answer. Further more, the variety and sheer quantity of the different types of tasks we may add to a production system can go up exponentially. This means that the task classification and plan retrieval phase may become much more complex and less reliable than the current test case. This may imply that SLM of a larger size may be needed to retain the performance and quality of service seen in the current analysis.

\section{Conclusion}\label{sec:conclusion}

We introduced the Nano Bio-Agent (NBA) framework, demonstrating that Small Language Models with fewer than 10 billion parameters can achieve performance comparable to much larger models in genomics question answering when deployed within an agentic architecture. This work validates the principle that architectural intelligence can overcome computational limitations without sacrificing accuracy.

Our evaluation across 50 models reveals that NBA enables SLMs to maintain robust performance across diverse genomics tasks while achieving significant efficiency improvements compared to conventional large model approaches. The framework shows consistent effectiveness across model families and genomics domains, from gene nomenclature to sequence alignment, indicating that task decomposition and tool orchestration successfully address the inherent limitations of smaller models in specialized domains.

These findings have promising implications for democratizing access to AI-powered genomics tools. By eliminating the requirement for expensive, large-scale models, NBA makes sophisticated genomics applications accessible to resource-constrained academic institutions, clinical environments, and educational settings. What's more, the data security concerns associated with private genomic data can be mitigated by ensuring that only on-premise SLMs handle sensitive information, avoiding unnecessary data sharing with external LLM vendors. This approach aligns with growing emphasis on sustainable ML development and supports the broader adoption of computational genomics across diverse research communities. 

Potential future directions in the short term include uncertainty quantification for reliability assessment, reinforcement learning fine-tuning for improved performance, and Model Context Protocol (MCP) integrations for enhanced interoperability. Additionally, we anticipate substantial opportunities in expanding the framework's tool repertoire and broadening its applicability to diverse tasks that would benefit from automated workflows.

The proposed modular design involving SLM positions the approach for extension to other scientific and business domains where specialized knowledge and tool integration are critical. As the field moves toward safer, more efficient and environmentally conscious language model deployment, our work provides a practical pathway for achieving state-of-the-art performance through architectural innovation rather than computational scaling.

\clearpage

\bibliographystyle{unsrt} 
\bibliography{bibliography}

\begin{thebibliography}{10}

\bibitem{sarumi2024large}
Oluwafemi~A. Sarumi and Dominik Heider.
\newblock Large language models and their applications in bioinformatics.
\newblock {\em Computational and Structural Biotechnology Journal},
  23:3498--3505, December 2024.
\newblock Mini-Review.

\bibitem{kalai2025languagemodelshallucinate}
Adam~Tauman Kalai, Ofir Nachum, Santosh~S. Vempala, and Edwin Zhang.
\newblock Why language models hallucinate, 2025.

\bibitem{ji2023survey}
Ziwei Ji, Nayeon Lee, Rita Frieske, Tiezheng Yu, Dan Su, Yan Xu, Etsuko Ishii,
  Yejin Bang, Andrea Madotto, and Pascale Fung.
\newblock Survey of hallucination in natural language generation.
\newblock {\em ACM Computing Surveys}, 55(12):1--38, 2023.

\bibitem{gao2023palprogramaidedlanguagemodels}
Luyu Gao, Aman Madaan, Shuyan Zhou, Uri Alon, Pengfei Liu, Yiming Yang, Jamie
  Callan, and Graham Neubig.
\newblock Pal: Program-aided language models, 2023.

\bibitem{mialon2023augmented}
Gr{\'e}goire Mialon, Roberto Dess{\`\i}, Maria Lomeli, Christoforos Nalmpantis,
  Ram Pasunuru, Roberta Raileanu, Baptiste Rozi{\`e}re, Timo Schick, Jane
  Dwivedi-Yu, Asli Celikyilmaz, et~al.
\newblock Augmented language models: a survey.
\newblock {\em arXiv preprint arXiv:2302.07842}, 2023.

\bibitem{jin2024genegpt}
Qiao Jin, Yifan Yang, Qingyu Chen, and Zhiyong Lu.
\newblock Genegpt: augmenting large language models with domain tools for
  improved access to biomedical information.
\newblock {\em Bioinformatics}, 40(2):btae075, 2024.

\bibitem{hou2023geneturing}
Wenpin Hou, Xinyi Shang, and Zhicheng Ji.
\newblock Benchmarking large language models for genomic knowledge with
  geneturing.
\newblock {\em bioRxiv}, pages 2023--03, 2023.

\bibitem{sayers2019database}
Eric~W Sayers, Richa Agarwala, Evan~E Bolton, J~Rodney Brister, Kathi Canese,
  Karen Clark, Rich Connor, Nicolas Fiorini, Kathryn Funk, Timothy Hefferon,
  et~al.
\newblock Database resources of the national center for biotechnology
  information.
\newblock {\em Nucleic acids research}, 47(D1):D23--D28, 2019.

\bibitem{altschul1990basic}
Stephen~F Altschul, Warren Gish, Webb Miller, Eugene~W Myers, and David~J
  Lipman.
\newblock Basic local alignment search tool.
\newblock {\em Journal of molecular biology}, 215(3):403--410, 1990.

\bibitem{chen2021evaluating}
Mark Chen, Jerry Tworek, Heewoo Jun, Qiming Yuan, Henrique Pond{\'e}
  de~Oliveira Pinto, Jared Kaplan, Harri Edwards, Yuri Burda, Nicholas Joseph,
  Greg Brockman, et~al.
\newblock Evaluating large language models trained on code.
\newblock {\em arXiv preprint arXiv:2107.03374}, 2021.

\bibitem{brown2020language}
Tom Brown, Benjamin Mann, Nick Ryder, Melanie Subbiah, Jared~D Kaplan, Prafulla
  Dhariwal, Arvind Neelakantan, Pranav Shyam, Girish Sastry, Amanda Askell,
  et~al.
\newblock Language models are few-shot learners.
\newblock {\em Advances in Neural Information Processing Systems},
  33:1877--1901, 2020.

\bibitem{belcak2025small}
Peter Belcak, Greg Heinrich, Shizhe Diao, Yonggan Fu, Xin Dong, Saurav
  Muralidharan, Yingyan~Celine Lin, and Pavlo Molchanov.
\newblock Small language models are the future of agentic ai, 2025.

\bibitem{lin2025bridging}
Anqi Lin, Junpu Ye, Chang Qi, Lingxuan Zhu, Weiming Mou, Wenyi Gan, Dongqiang
  Zeng, Bufu Tang, Mingjia Xiao, Guangdi Chu, et~al.
\newblock Bridging artificial intelligence and biological sciences: a
  comprehensive review of large language models in bioinformatics.
\newblock {\em Briefings in Bioinformatics}, 26(4):bbaf357, 2025.

\bibitem{su2025biomaster}
Houcheng Su, Weicai Long, and Yanlin Zhang.
\newblock Biomaster: Multi-agent system for automated bioinformatics analysis
  workflow.
\newblock {\em bioRxiv}, pages 2025--01, 2025.

\bibitem{wang2025geneagent}
Zhizheng Wang, Qiao Jin, Chih-Hsuan Wei, Shubo Tian, Po-Ting Lai, Qingqing Zhu,
  Chi-Ping Day, Christina Ross, Robert Leaman, and Zhiyong Lu.
\newblock Geneagent: self-verification language agent for gene-set analysis
  using domain databases.
\newblock {\em Nature Methods}, 22:1677--1685, 2025.

\bibitem{subramanian2025smalllanguagemodelsslms}
Shreyas Subramanian, Vikram Elango, and Mecit Gungor.
\newblock Small language models (slms) can still pack a punch: A survey, 2025.

\bibitem{garg2025risesmalllanguagemodels}
Muskan Garg, Shaina Raza, Shebuti Rayana, Xingyi Liu, and Sunghwan Sohn.
\newblock The rise of small language models in healthcare: A comprehensive
  survey, 2025.

\bibitem{gridach2025agentic}
Mourad Gridach, Jay Nanavati, Khaldoun Zine~El Abidine, Lenon Mendes, and
  Christina Mack.
\newblock Agentic ai for scientific discovery: A survey of progress,
  challenges, and future directions.
\newblock In {\em International Conference on Learning Representations (ICLR)}.
  IQVIA, 2025.

\bibitem{alphagenome}
{\v Z}iga Avsec, Natasha Latysheva, Jun Cheng, Guido Novati, Kyle~R. Taylor,
  Tom Ward, Clare Bycroft, Lauren Nicolaisen, Eirini Arvaniti, Joshua Pan,
  Raina Thomas, Vincent Dutordoir, Matteo Perino, Soham De, Alexander Karollus,
  Adam Gayoso, Toby Sargeant, Anne Mottram, Lai~Hong Wong, Pavol Drot{\'a}r,
  Adam Kosiorek, Andrew Senior, Richard Tanburn, Taylor Applebaum, Souradeep
  Basu, Demis Hassabis, and Pushmeet Kohli.
\newblock {AlphaGenome}: advancing regulatory variant effect prediction with a
  unified {DNA} sequence model.
\newblock {\em bioRxiv}, 2025.

\end{thebibliography}

\appendix

\section{Appendix / Supplementary Material} \label{sec:appendix} 
\subsection{Code Availability and Reproducibility}

The complete implementation of the NBA framework is publicly available in the GitHub repository: \url{https://github.com/georgesshong/nanobioagent}. The repository includes all source code, configuration files, and documentation necessary to reproduce the experiments presented in this paper.

All experiments were conducted using Python 3.9 with the following key dependencies:
\begin{itemize}
\item LangChain 0.3.24 for agentic framework implementation
\item NumPy 2.0.2 and Pandas 1.5+ for data processing
\item Anthropic 0.50.0, OpenAI 1.76.0 for API access
\item Hugging Face Hub 0.30.0 for model loading
\end{itemize}
A complete requirements.txt file with exact versions is provided in the repository. Evaluation scripts reproduce all figures and tables presented in this paper. The repository also contains configuration files for all 50 language models tested.

\subsection{Refinement of Ground Truth Dataset} \label{sec:appendix_refinement_dataset} 

To ensure accurate benchmarking, the original \href{https://github.com/Winnie09/GeneTuring}{GeneTuring} question/answer dataset from the \href{https://github.com/ncbi/GeneGPT}{GeneGPT} GitHub was updated to rectify discrepancies identified during manual verification against the NCBI and Ensembl databases.

A significant issue concerned questions about human genome DNA alignment, a limitation acknowledged by the original authors in the paper. They noted that the lack of a specified reference genome made precise coordinate validation impossible, leading them to adopt a partial scoring system. As they state in their paper, "For the DNA sequence alignment to human genome task, we give correct chromosome mapping but incorrect position mapping a score of 0.5... since the original task does not specify a reference genome" ~\cite{jin2024genegpt}. This 0.5 score created an artificial performance ceiling that misrepresented the true capabilities of the evaluated models. To address this, we developed a programmatic solution to query and validate the precise genomic coordinates against a defined reference, establishing a new, verifiable ground truth that was applied consistently across all methods.

Additional minor updates were implemented to correct outdated or inaccurate information. For example, some questions referenced invalid identifiers, such as \texttt{ENSG10010137820.1}, which does not exist in the official Ensembl database for \textit{Homo sapiens} (GRCh38)\footnote{\url{https://www.ensembl.org/biomart/martview/}}. Other corrections involved aligning answers with the current NCBI knowledgebase; for instance, acceptable answers to the query for the official symbol of "PTH1" is correctly updated to include \href{https://www.ncbi.nlm.nih.gov/datasets/gene/138428/}{\textcolor{blue}{\texttt{PTRH1}}}. In total, only 2\% of the questions containing such irreconcilable disagreements were excluded from the scoring. The resulting ground truth dataset is fully robust and programmatically aligned with information retrieved from current NCBI records. A comprehensive log of all modifications, including our updated ground truth annotations, is available in the project's GitHub repository.

\end{document}